\documentclass[runningheads]{llncs}
\usepackage{graphicx}
\usepackage{amsmath,amssymb} 
\usepackage{mathtools}
\usepackage{color}
\usepackage[inline]{enumitem}
\usepackage{siunitx}
\usepackage[nolist,nohyperlinks]{acronym}
\usepackage{booktabs}

\begin{document}
\pagestyle{headings}
\mainmatter

\def\ACCV18SubNumber{2}  

\newcommand{\etal}{et al.}

\DeclarePairedDelimiter\abs{\lvert}{\rvert}%

\title{LoANs: Weakly Supervised Object Detection with Localizer Assessor Networks} 
\titlerunning{LoANs: Localizer Assessor Networks}
\authorrunning{Bartz, Yang, Bethge, Meinel}

\author{Christian Bartz, Haojin Yang, Joseph Bethge \and Christoph Meinel \\ \{christian.bartz, haojin.yang, joseph.bethge, christoph.meinel\}@hpi.de}
\institute{Hasso Plattner Institute, University of Potsdam \\ Prof.-Dr.-Helmert Strasse 2-3 \\ 14482 Potsdam, Germany}

\maketitle

\begin{acronym}
	\acro{CNN}{Convolutional Neural Network}
	\acro{DNN}{Deep Neural Network}
	\acro{IOU}{Intersection over Union}
	\acro{GAN}{Generative Adversarial Network}
\end{acronym}

\begin{abstract}
Recently, deep neural networks have achieved remarkable performance on the task of object detection and recognition.
The reason for this success is mainly grounded in the availability of large scale, fully annotated datasets, but the creation of such a dataset is a complicated and costly task.
In this paper, we propose a novel method for weakly supervised object detection that simplifies the process of gathering data for training an object detector.
We train an ensemble of two models that work together in a student-teacher fashion.
Our student (localizer) is a model that learns to localize an object, the teacher (assessor) assesses the quality of the localization and provides feedback to the student.
The student uses this feedback to learn how to localize objects and is thus entirely supervised by the teacher, as we are using no labels for training the localizer.
In our experiments, we show that our model is very robust to noise and reaches competitive performance compared to a state-of-the-art fully supervised approach.
We also show the simplicity of creating a new dataset, based on a few videos (e.g. downloaded from YouTube) and artificially generated data.
\end{abstract}

\section{Introduction}

One of the main factors for the success of \acp{DNN}, in the recent years~\cite{Krizhevsky2012Imagenet,He2016Deep,Ren2015Faster}, is the availability of large-scale labeled datasets like the ImageNet dataset~\cite{Deng2009Imagenet}.
In the domain of object detection, fully annotated datasets like Pascal VOC~\cite{Everingham2010Pascal} enabled several breakthroughs for object detectors~\cite{Ren2015Faster,Redmon2016You,Liu2016Ssd}.
These methods heavily rely on annotated bounding boxes for each object in an image.
Semi-supervised / weakly supervised methods for object detection~\cite{Deselaers2010Localizing,Liang2015Towards,Misra2015Watch,Tang2016Large} try to overcome the high costs of labeling by using less annotations.
We review these and more methods related to our work in section~\ref{sec:related_work}.
All mentioned methods have in common that they do need some form of annotation for each input image.
Creating these annotated datasets incurs a high amount of manual labor that has to be performed in order to label the data and make it available for creating a computer vision model.
Having the possibility to get enough annotated training data for a specialized application without the high annotation costs would be ideal.
To this end, we propose a novel approach for weakly supervised object detection.
We use an ensemble of two independent neural networks that are jointly trained.
The first network (localizer) is trained by the second network and learns to perform the task of object localization in a given input image.
The second network (assessor) is trained to regress the \ac{IOU} (also known as Jaccard Index) of the bounding box of an object and an image crop.
The assessor is trained in a fully supervised way on purely artificially generated data.
We describe the architecture of our system in more detail in section~\ref{sec:proposed_system}.
The basic data necessary for generating the training set for the assessor consists of a few template images of the objects that shall be localized and a few different natural background images.
In our experiments, we used 25 template images and 8 background images to train an assessor for the task of localizing figure skaters.
The localizer, on the contrary, does not need any annotations, as it is entirely trained by the supervision of the assessor.
The data used for training the localizer could, for instance, come from a 5 to 10 minute long video that has been downloaded from the internet, or especially created for this task.
We provide further information about the datasets we used and also the experiments we performed in section~\ref{sec:experiments} and conclude our work in section~\ref{sec:discussion_and_future_work}.

The contributions of our work can be summarized as follows:
\begin{enumerate*}[label={(\arabic*)}]
	\item we propose a novel (end-to-end) training method for weakly supervised object detection, based on knowledge transfer between two jointly trained, but independent neural networks.
	\item Our proposed model can successfully be trained, even if more than \SI{50}{\percent} of the images in the train dataset are noisy images (i.e. images that do not contain an object we are looking for, or contain only a partial view of the object).
	\item We show that our model reaches competitive performance compared to a state-of-the-art fully supervised object detection system.
	\item In our experiments, we show that short video clips, plus a few template and background images that were gathered from the internet, are sufficient to create a new dataset and successfully train a model.
	\item We release our code, our models, and the generated data sets to the community.\footnote{\url{https://github.com/Bartzi/loans}}
\end{enumerate*}

\section{Related Work}
\label{sec:related_work}

Object Detection has been intensively studied in the past years.
Thanks to the availability of huge amounts of labeled data it is possible to create fully supervised systems that achieve incredible results for object detection and recognition~\cite{Ren2015Faster,Redmon2016You,Liu2016Ssd,Girshick2015Fast,Singh2018Sniper}.
Those methods are fully-supervised, meaning that they need annotations for the location of objects (bounding boxes) and also annotations for the class of each object in the image.
Getting fully annotated images is a costly process, especially for applications beyond academic use cases.

\subsubsection{Weakly Supervised Localization}
Weakly supervised object detection systems try to overcome the annotation problem, by learning to detect objects with partial bounding box annotations or even without the need for this kind of annotations~\cite{Deselaers2010Localizing,Liang2015Towards,Misra2015Watch,Tang2016Large}.
The approach introduced by Deselaers~\etal~\cite{Deselaers2010Localizing} uses object detectors already trained on certain classes to annotate the locations of objects in new classes.
Those labeled images can then be used to train a fully supervised detector on the new class.
Liang~\etal~\cite{Liang2015Towards} propose a method that adapts a pre-trained classification model for object detection on new and unseen classes, while needing only a few fully annotated instances and videos that are likely to contain objects of the new class.
Misra~\etal~\cite{Misra2015Watch} follow a similar approach by starting with a small set of annotated bounding boxes and iteratively extracting more annotations from unannotated video sequences.
Tang \etal~\cite{Tang2016Large} propose a system that uses information from fully annotated visually and semantically similar classes to train an object detector on classes that are only partially annotated.
Most of these approaches rely on images that at least have a category label, which is in contrast to our proposed method.
For training the localizer, we do not need any labels at all, as the localizer is trained by the assessor.
The assessor, on the contrary, only needs annotations in form of template images that are placed randomly in natural images.

\subsubsection{Knowledge Transfer between Neural Networks}
Knowledge transfer has also been intensively studied in the recent years~\cite{Hinton2015Distilling,Tang2016Recurrent,Xie2016Cooperative,Chen2016Net2Net,Li2017Learning,Jiang2018Mentornet}.
Existing work mostly concentrates on using one network (the teacher) to teach another network (the student), to perform the same task, but either increasing the performance of the model~\cite{Chen2016Net2Net,Li2017Learning}, or compressing the model, while keeping the same level of performance~\cite{Hinton2015Distilling,Tang2016Recurrent}.
Those approaches use a pre-trained model and try to distill or adapt this model into the other.
Other approaches train both models at the same time.
Jiang~\etal~\cite{Jiang2018Mentornet} make the teacher network behave like a `real' teacher that provides a curriculum learning strategy for the student.
This curriculum helps the student to successfully learn to perform its task on noisy data.
In this setting, knowledge transfer is also used for training an image generator, where the teacher acts as an advisor that refines the images generated by the student, while getting additional feeback from the outside world~\cite{Xie2016Cooperative}.
Apart from the approach by Jiang~\etal~\cite{Jiang2018Mentornet}, all knowledge transfer approaches have in common that teacher and student deal with problems from the same domain (i.e. computing the same function, or teacher and student work on image generation).
In our proposed method, teacher and student (or assessor and localizer) work in different problem domains.
Our assessor is a model that is used to predict the \ac{IOU} between an image crop and the bounding box of an object in this crop, while the localizer is trained to find and crop objects from an image, by using the feedback of the assessor.

\subsubsection{Generative Adversarial Networks}
Our approach is, to some extent, inspired by \acp{GAN} that have been introduced by Goodfellow~\etal~\cite{Goodfellow2014Generative}.
In a \ac{GAN}, two networks (generator and discriminator) are trained simultaneously, and the goal of both networks is to work against each other.
The generator is trying to produce images that are indistinguishable from real-world images.
At the same time, the discriminator is trying to decide whether an analyzed sample is a real sample or has been generated by the generator.
In our method, the assessor (discriminator in the \ac{GAN} setting) and localizer (generator in the \ac{GAN} setting) are two independent networks that are trained simultaneously.
The objective of the localizer is not to fool the assessor, but to maximize the output of the assessor.
This leads to the situation that the localizer is trained by the supervision of the assessor, while the assessor does not even know that it is used to train another neural network.

\section{Proposed System}
\label{sec:proposed_system}

When humans first see a new object they memorize certain aspects of the object and create a template that is matched against new occurences of the same object class.
We mimic a similar behavior, by creating a system that can be trained to localize an object, using only a few template images of the object, which are placed randomly in a natural image and also unlabeled images that are likely to contain the object, which is to be detected (we refer to this kind of object as the ``target object'' for the remainder of this paper).
Our proposed system consists of two independent \acp{DNN} (localizer and assessor) that are trained at the same time.
In this section, we start with explaining the assessor, followed by an explanation of the localizer, and how both networks are jointly trained.

\subsection{Assessor}

The first network is the assessor.
The assessor receives an input image $I_A$ and produces a value $y \in [0,1]$.
The value $y$ provides a measure of the ratio of an object contained in the image $I_A$.
In other words, the assessor predicts the \ac{IOU} of the bounding box of an object and the input image.
The prediction of this \ac{IOU} is trained in a fully-supervised fashion based on purely artificially generated data.
The assessor consists of a \ac{DNN} that computes a function $f_{as}(I_A)$.
The function computed by the assessor is defined as follows:
\begin{equation}
	f_{as}(I_A) = y = \sigma(f_{dnn}(I_A))~.
\end{equation}
$f_{dnn}(I_A)$ is a \ac{DNN} that produces a scalar value, and $\sigma$ denotes the logistic sigmoid function, which forces the values of $y$ to be in the interval $[0,1]$.
While training, the assessor minimizes the mean squared error between the prediction $y$ and the groundtruth label $l \in [0,1]$ ($n$ denotes the batch size used during training): 
\begin{equation}
	\label{eq:assessor_cost}
	\mathcal{L}_{assessor} = \frac{1}{n} \sum_{i=1}^n (f_{as}(I_A^i) - l^i)^2~.
\end{equation}

\subsubsection{Data Generation}

\begin{figure}[tb]
	\centering
	\includegraphics[width=0.95\textwidth]{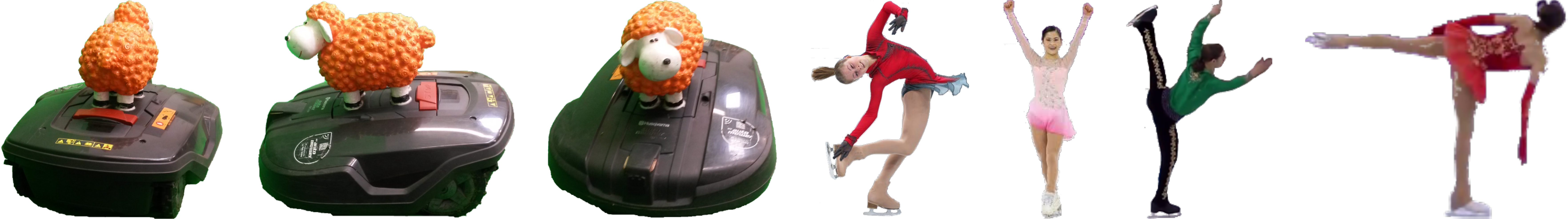}
	\caption{
		Template images of objects used for creating the train dataset of the assessor.
		These template images have been created by cropping the object from real images. 
	}
	\label{fig:assessor_dataset_template}
\end{figure}

\begin{figure}[b]
	\centering
	\includegraphics[width=0.95\textwidth]{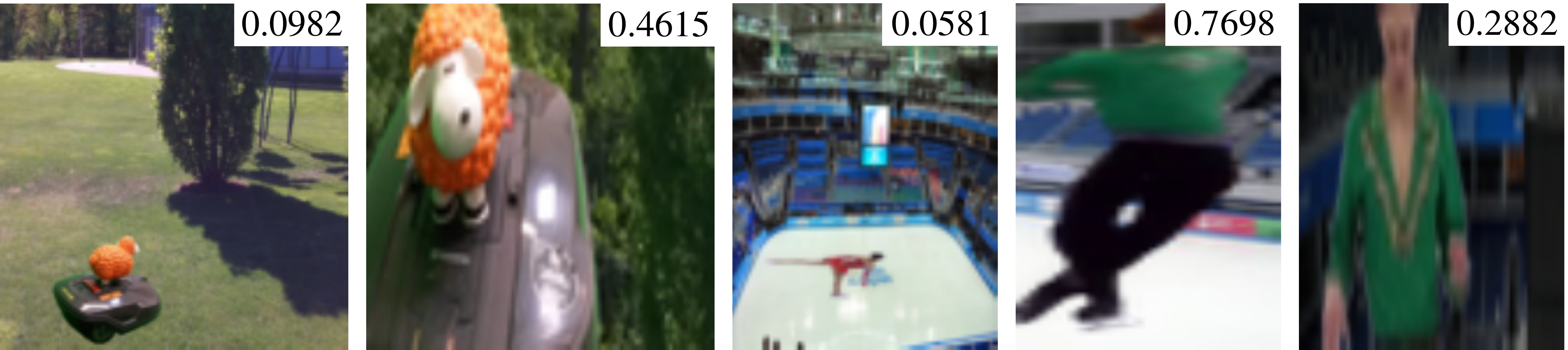}
	\caption{
		Sample input images and their corresponding labels for the assessor.
		Each image contains some portion of an object.
		The label is the \acl{IOU} of the image with the bounding box of the object. 
		In other words, it is the proportion of the object that covers the image.
	}
	\label{fig:assessor_input_example}
\end{figure}

The assessor is trained on an entirely synthetic dataset.
This dataset can be created by using some template images of the target object.
For us, a template is an object that has been cropped from a natural image and can be pasted into other natural images.
Figure~\ref{fig:assessor_dataset_template} shows some examples of template images that we used for our experiments.
The number of template images that are necessary to create a good model depends on the number of pose variants the type of object may have.
Besides the template images of the target objects, we also need some background images that do not contain the target object.
The template images are pasted onto the background images at random locations, with random sizes.
While pasting template images onto the background images, we do not care that some images might be placed at locations where they might never occur in a real world setting.
Figure~\ref{fig:assessor_input_example} shows some generated images that are used for training the assessor.

\subsection{Localizer}

The second network is the localizer.
The localizer receives a natural image as input and tries to crop a region from the input image that is likely to contain an object. 
The localizer uses a spatial transformer~\cite{Jaderberg2015Spatial} to crop the region from the image.
A spatial transformer is a differentiable module for \acp{DNN} that applies a spatial transformation on an input feature map $I_L$ and produces an output feature map $O$.
Such a spatial transformer module consists of three different parts: Localization Network, Grid Generator, and Image Sampler.

\subsubsection{Localization Network}
The localization network is a neural network that computes a set of parameters $\theta$.
These parameters define the spatial transformation that is to be applied on the input feature map.
These parameters can describe different types of spatial transformations, such as affine or thin plate spline transformations.
In our work, we only use affine transformations that allow us to attend to different regions of the input image $I_L$, by using the transformation to zoom into a region of interest.
The localization network learns a function $g_{dnn}(I_L)$ that predicts the parameters $\theta_i$ of an affine transformation matrix $A_{\theta}$ that is conditioned on the input image $I_L$ and the parameters of the neural network.
The result of this function is defined as follows:
\begin{equation}
	g_{dnn}(I_L) = A_{\theta} = 
	\begin{bmatrix}
		\theta_1 & 0 & \theta_2 \\
		0 & \theta_3 & \theta_4 \\
	\end{bmatrix}~.
\end{equation}
$\theta_1$ to $\theta_4$ are the transformation parameters, regressed by the deep neural network.
We constrain the parameters of the affine transformation to only allow cropping, translation, and scaling.

\subsubsection{Grid Generator}
After the prediction of the affine transformation matrix, the grid generator is used to create a regularly spaced grid consisting of coordinates $x_{w_o}, y_{h_o}$, with height $H_o$ and width $W_o$ being the spatial size of the output feature map $O$.
The regularly spaced grid is used together with the already predicted transformation matrix to produce a regular grid $G$ of sampling coordinates $u_{i}, v_{j}$, with $i \in [0, \dots, W_o]$ and $j \in [0, \dots, H_o]$.
\begin{equation}
	\begin{pmatrix}
		u_{i} \\
		v_{j}
	\end{pmatrix}
	= A_{\theta}
	\begin{pmatrix}
		x_{w_o}	\\
		y_{h_o} \\
		1
	\end{pmatrix}
	= \begin{bmatrix}
		\theta_1 & 0 & \theta_2 \\
		0 & \theta_3 & \theta_4 \\
	\end{bmatrix}
	\begin{pmatrix}
		x_{w_o}	\\
		y_{h_o} \\
		1
	\end{pmatrix}~.
\end{equation}
We can use these coordinates to determine the bounding box of the region of interest, which we want to extract from the image.
This bounding box is the intended output of our localizer and is used to display the detection result of the network.

\subsubsection{Image Sampler}
In order to provide an input image to our assessor, the sampling grid produced by the grid generator is used to sample the input image $I_L$ at the sampling points $u_i, v_j$ of the generated sampling grid $G$.
As the generated sampling grid will not perfectly align with the values in the discrete grid of the input image, we use bilinear sampling and define the values of each pixel $i,j$ ($i \in [0, \dots, W_o]$ and $j \in [0, \dots, H_o]$) of the output image $O$ as:
\begin{equation}
	f_{loc}(I_L)_{ij} = O_{ij} = \sum^{W_o}_{w} \sum^{H_o}_h I_{L_{wh}}\, max(0, 1 - \abs{u_i - w})\, max(0, 1 - \abs{v_j - h})~.
\end{equation}

This formulation of bilinear sampling is (sub-)differentiable, thus allowing us to propagate error gradients from the assessor to the localization network, using standard backpropagation.
The output of the bilinear sampling operation is also the final output of the localizer that represents the function $f_{loc}(I_L)$.

The combination of the three modules \textit{Localization Network}, \textit{Grid Generator}, and \textit{Image Sampler} forms our localizer.
During training we extract the output image, produced by the \textit{Image Sampler} and hand it over to the assessor to assess the quality of the detection.

\subsubsection{Training Data}

We do not need annotated data for training the localizer, as the training of the localizer is supervised by the assessor.
This property of the localizer makes it very easy to generate a large scale database of training images.
One possibility for getting training data is to extract frames that might contain the target object from a video clip.
The frames can directly be used as input to the localizer and creating them does not include any manual labor, except from choosing appropriate videos.
Those videos could contain noisy frames, such as frames that do not show any target object, or only some parts of a target object.
In our experiments (section~\ref{sec:experiments}), we show that a noisy dataset does not necessarily degrade the performance of our approach, it might even improve the performance of our model on a test dataset.

\begin{figure}[tb]
	\centering
	\includegraphics[width=0.95\textwidth]{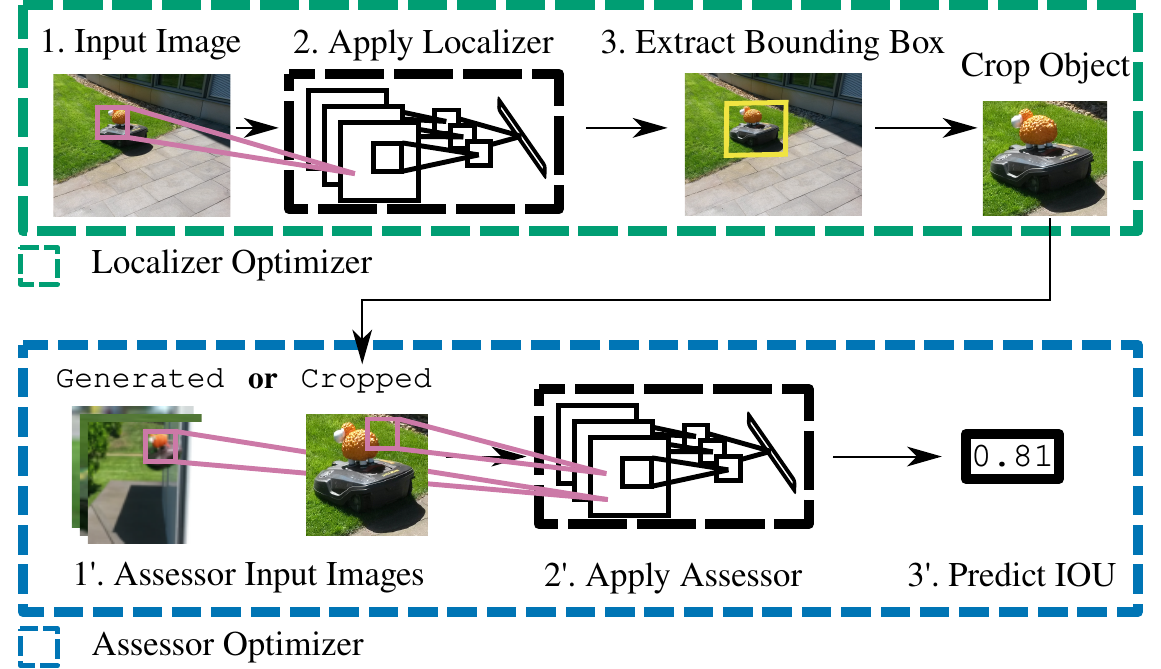}
	\caption{
		We use two networks for training our model, each with its own optimizer.
		The first network is the localizer and the second network is the assessor.
		The input to the localizer is an image containing the target object.
		The localizer extracts the bounding box of the localized object and also crops the object from the image and provides it to the assessor as an input.
		The assessor predicts the \acl{IOU} of the object and the image crop and is used by the localizer to learn its task.
		The striped lines indicate which parts of the two networks are handled by which optimizer.
	}
	\label{fig:structure_of_network}
\end{figure}

\subsection{Optimization of both Networks}
The two parts of our model, as such, are independent neural networks.
It is possible to train both networks at the same time, but they can also be trained consecutively.
It is important that the assessor is at some time able to provide meaningful feedback to the localizer, thus it does not make sense to train the localizer without an assessor that is either pretrained or is learning to regress the \ac{IOU} of an object and the image crop at the same time.
For our experiments (see section~\ref{sec:experiments}), we always trained both networks at the same time.

While training both networks at the same time, we use two independent optimizers.
One optimizer handles the weight updates of the assessor.
The second optimizer handles the weight updates of the localizer.
Figure~\ref{fig:structure_of_network} provides a structural overview of our proposed system.
The localizer learns to localize objects, by using the assessor to assess the quality of the object that has been localized by the localizer.
This means that the cost $\mathcal{L}$ of the localizer is the cost assigned by the assessor to the part of the image cropped by the localizer, plus two localizer specific regularizers:
\begin{equation}
	\mathcal{L}_{localizer} = \frac{1}{n} \sum_{i=1}^n (f_{as}(f_{loc}(I_L^i)) - 1.0)^2 + \mathcal{L}_{di}(G^i) + \mathcal{L}_{out}(G^i)~. \label{eq:loss_localizer}
\end{equation}
Where $\mathcal{L}_{di}(G)$ is a regularization term based on the direction of the predicted sampling grid $G$.
This regularization term penalizes grids that are mirrored along the x- or y-axis.
$\mathcal{L}_{out}(G)$ is a regularization term that penalizes predicted grids with coordinates that lie outside of the image coordinates.
The label for the mean squared error loss in equation \eqref{eq:loss_localizer} is constant for every input image.
We set this constant to $1$, as we want our localizer to crop regions of interest that fully contain the target object with as little extra space as possible.
It would also be possible to set this value to any other value in the interval $[0,1]$.
Setting this value to $0$, for instance, results in a localizer that deliberately chooses regions without the target object.

The most important part of training both networks at the same time, is the actual weight update process.
Once we obtain the cost for the assessor (as described in equation \eqref{eq:assessor_cost}), we calculate gradients using standard backpropagation and update the weights of the assessor.
Updating weights for the localizer works in a similar fashion.
Here, we also calculate the gradients with backpropagation and let the gradients flow through the assessor to the localizer and to the first neural network layer.
Once we obtained the gradients, we only update the weights of the localizer and leave the weights of the assessor untouched, hence the localizer is trained by the assessor to find a good region of interest.

\section{Experiments}
\label{sec:experiments}

In this section we evaluate our proposed system on two different real-world datasets.
We performed our experiments, in order to answer the following questions:
\begin{enumerate*}[label={(\arabic*)}]
	\item can we use the error information obtained by one network to train a second network on a different task, while the first network does not know that it is used in such a way?
	\item Is it feasible to train the assessor on totally synthetic data, where we do not care whether the samples are realistic or not, and at the same time use unlabeled data for training an object localizer?
	\item Can we hold our promise that it is simple to create a new dataset, using our proposed approach?
	\item Is our model able to reach competitive performance, compared to a fully supervised model?
\end{enumerate*}

In order to answer these questions, we performed experiments on two different datasets, we created ourselves.
The first dataset, contains images of an automated lawn-mower that has an orange sheep on top.
The second dataset contains images of figure skaters giving a performance, for instance at the Olympic Games.

We begin this section, by introducing the datasets we used throughout our experiments in detail.
This description is followed by an explanation of our experimental setup.
We conclude the section with showing and discussing the results of our experiments.

\subsection{Datasets}
\label{subsec:datasets}

We created two challenging real-world datasets, which we tested our method on.
In the first dataset we are trying to localize an automated lawn-mower that has an orange sheep on top.
This dataset is challenging, because it contains patches where the sheep is only a few pixels high and wide.
We refer to this object for the remainder of the paper as ``sheep''.
In the second dataset we tried to localize figure skaters while they are giving a performance.
The figure skater dataset contains a lot of pose variations for each of the skaters, making it challenging for a model to generalize.

\subsubsection{Sheep Dataset}
\label{subsubsec:sheep_dataset}
We created this dataset by taking \num{158} different images that we used as background images for the assessor dataset.
Besides the backgrounds, we took \num{10} pictures of the sheep from different angles that are used as template images.
We used \num{8} template images to randomly paste them onto the background images for the assessor dataset.
Using this data, we generated \num{10000} different images for training the assessor.

Out of the \num{158} backgrounds, we used \num{121} randomly chosen backgrounds and the same \num{8} template images to create a train dataset for the localizer.
We pasted the template images onto the background images using positions of hand-crafted bounding boxes, rendering those images as similar to real-world images as possible.
We did not use the bounding box information for training the model with our approach, but for training the model with the fully supervised baseline approach.
All in all, we have been able to generate \num{8320} images for training the localizer.

We used the remaining \num{37} backgrounds and the last two template images to create a test dataset where we also pasted the sheep onto the background images using hand-crafted bounding boxes.
Using this approach, we were able to obtain \num{560} images for testing the localizer.

\subsubsection{Figure Skating Dataset}
\label{subsubsec:figure_skating_dataset}

In order to create this dataset we downloaded \num{5} videos from Youtube, downloaded \num{8} background images, and created \num{25} template images.
Using the \num{8} background images and \num{25} template images, we generated \num{10000} images for training the assessor.

For the train dataset of the localizer, we took two videos from the Olympic Games in Pyeong Chang 2018 with the performance of Alina Zagitova\footnote{\url{https://www.youtube.com/watch?v=TlXCk1LDlC0}} and Yuzuru Hanyu.\footnote{\url{https://www.youtube.com/watch?v=23EfsN7vEOA}}
We took one video from the Olympic Games of Sochi in 2014 with the performance of Yulia Lipnitskaya.\footnote{\url{https://www.youtube.com/watch?v=ke0iusvydl8}}
Last, we took the performance of Jason Brown at the US Open in 2014.\footnote{\url{https://www.youtube.com/watch?v=J61k2XjRryM}}
After extracting all frames from the videos we were left with \num{68985} images for training the localizer.
As these images still contain a lot of noise, where the figure skater is either not entirely, or not at all present in an image, we created a second train dataset without images containing this kind of noise.
The second, noise free, dataset contains \num{33125} images, which is roughly \SI{48}{\percent} of the original number of images.
For testing the localizer, we used the performance of Yuna Kim at the Olympic Games in Sochi 2014.\footnote{\url{https://www.youtube.com/watch?v=hgXKJvTVW9g}}
We extracted \num{100} images from this video that shows Yuna Kim in different positions and manually annotated them with bounding boxes for testing our model.

We were able to very quickly generate the train datasets, for both assessor and localizer, as it only took roughly \SI{1.5}{\hour}.
This shows that our proposed system makes it possible to create a new dataset for object localization with minimal effort.

\subsection{Experimental Setup}

In the following we explain the exact configuration of the neural network achitectures that we used during our experiments and also provide some implementation details.

\subsubsection{Localizer}

We use two different network architectures for training our localizer (ResNet-18 and ResNet-50).
Both architectures are based on the ResNet architecture proposed by He~\etal~\cite{He2016Deep}.
The input to the localizer is the image where the target object shall be localized.
Before passing the image to the network, we normalize the image, by subtracting each channel with a mean value obtained from the Imagenet dataset.
Before the first residual block of the network, we perform a $3 \times 3$ convolution with \num{64} output channels and stride \num{1}, followed by batch normalization~\cite{Ioffe2015Batch}, ReLU~\cite{Nair2010Rectified}, and a $3 \times 3$ max pooling layer with stride \num{2}.
After these layers, \num{4} residual blocks follow.
Each block consists of at least \num{6} $3 \times 3$ convolutional layers with stride \num{1}, the first convolutional layer of the second, third, and fourth residual block uses a stride of \num{2}. Each convolutional layer is followed by batch normalization and ReLU.
The number of convolutional filters is $64, 128, 256, 512$, or $256, 512, 1024, 2048$ for ResNet-18 and ResNet-50, respectively.
We perform global average pooling after the last residual block.
The last residual feature map, is fed to a fully connected layer with \num{6} neurons that predicts the affine transformation parameters that are used to generate the sampling grid.
We always set the parameters that are responsible for rotating the input image to zero.

\subsubsection{Assessor}
We based the architecture of the assessor network on the ResNet architecture, too.
We explicitly chose the ResNet architecture in order to mitigate the vanishing gradient problem and keep a strong gradient throughout both of our networks.
The input to the assessor is an image that resembles a crop from a larger image, which contains an object.
The assessor consists of four residual blocks, where the first and second block consist of three convolutional layers.
The first layer, is a $3 \times 3$ convolutional layer.
The last two layers are $4 \times 4$ convolutional layers with a stride of \num{2}.
The second layer is followed by the ReLU activation function.
The last two residual blocks consist of two $3 \times 3$ convolutional layers, followed by the ReLU activation function, each.
The number of convolutional feature maps for each residual block is $128, 128, 128, 128$, respectively.
A fully connected layer with \num{1} neuron and the sigmoid activation function $\sigma$ follows after the residual blocks.
We do not use a bias term for any of the layers in the assessor.

\subsubsection{Hyperparameters}
For all of our experiments, we use Adam~\cite{Kingma2015Adam} as optimizer for both localizer and assessor and set the learning rate (alpha) to $10^{-4}$.
We use a batch size of \num{32} for each experiment, and let the training run for \num{300} epochs.
We train the localizer with data augmentation.
We apply a random combination of horizontal flipping, color jittering, and random cropping / padding to \SI{50}{\percent} of the images in each batch.

\subsubsection{Implementation}
We implemented all of our experiments using Chainer~\cite{Tokui2015Chainer} and ChainerCV~\cite{Niitani2017Chainercv}.
We used one NVIDIA 1080Ti GPU for each experiment.

\subsubsection{Evaluation Metrics}
We follow the evaluation procedure of the Pascal VOC 2012 challenge~\cite{Everingham????Pascal}, where we calculate the average precision, based on the \ac{IOU} of the predicted bounding box and the ground truth bounding box.

\begin{table}[tb]
	\caption{
		Results of our experiments on the sheep dataset.
		We show the average precision for each model and the respective input sizes.
		The first row (SSD) shows the results of our trained baseline model.
		The other rows show the results of our models based on different ResNet architectures.
	}
	\begin{center}
		\setlength{\tabcolsep}{6pt}
		\begin{tabular}{lccc}
			\toprule
			Method & $224 \times 224$ & $300 \times 300$ & $512 \times 512$ \\
			\midrule
			SSD~\cite{Liu2016Ssd} & - & 0.887 & 0.969 \\
			ResNet-18 & 0.887  & 0.937 & 0.967 \\
			ResNet-50 & 0.959 & 0.958 & 0.976 \\
			\bottomrule
		\end{tabular}
	\end{center}
	\label{tab:sheep_experiments}
\end{table}

\subsection{Sheep Experiments}

We performed our first experiments on the sheep dataset, which we introduced in section~\ref{subsubsec:sheep_dataset}, to prove that our localizer-assessor concept works.
Since we have a fully annotated dataset for training and testing the localizer, we trained a baseline model following the SSD approach of Liu~\etal~\cite{Liu2016Ssd}, using different input sizes, i.e. images with $300 \times 300$ and $512 \times 512$ pixels.
We then used the same dataset (without bounding box annotations) for training the localizer, and the dataset we created for the assessor to train different models, based on our approach.
We trained models, with different input sizes ($224 \times 224$, $300 \times 300$, and $512 \times 512$ pixels), using different network architectures for the localizer (ResNet-18, or ResNet-50), and with an output size of $75 \times 75$ pixels for the localizer.
We always initialized each ResNet model with convolutional layers that have been pre-trained on the Imagenet dataset.
Table~\ref{tab:sheep_experiments} shows the results of our different experiments.
We also show some samples from the dataset including the predictions of our localizer in the left column of figure~\ref{fig:example_predictions}.
In our supplementary material, we provide a real-life video that shows the performance of our best model.
From the results in table~\ref{tab:sheep_experiments} we can clearly see that our model reaches competitive performance compared to a fully supervised model that was trained on the same dataset.
This is remarkable, as we do not use any labels for the location of the sheep while training our model.
We note, however, that our approach is currently not able to work with multiple objects in one image and also only supports the localization of objects of one class and not multiple classes, yet.
Another interesting observation is that our ResNet-18 based model significantly increases its localization performance across the different input sizes, while the ResNet-50 based model only does so in a smaller margin. 
We think that this is because the ResNet-50 model has a better representational capability, because it is deeper and that this is already enough to extract meaningful features, even at a low spatial resolution.

\begin{figure}[tb]
	\centering
	\includegraphics[width=0.95\textwidth]{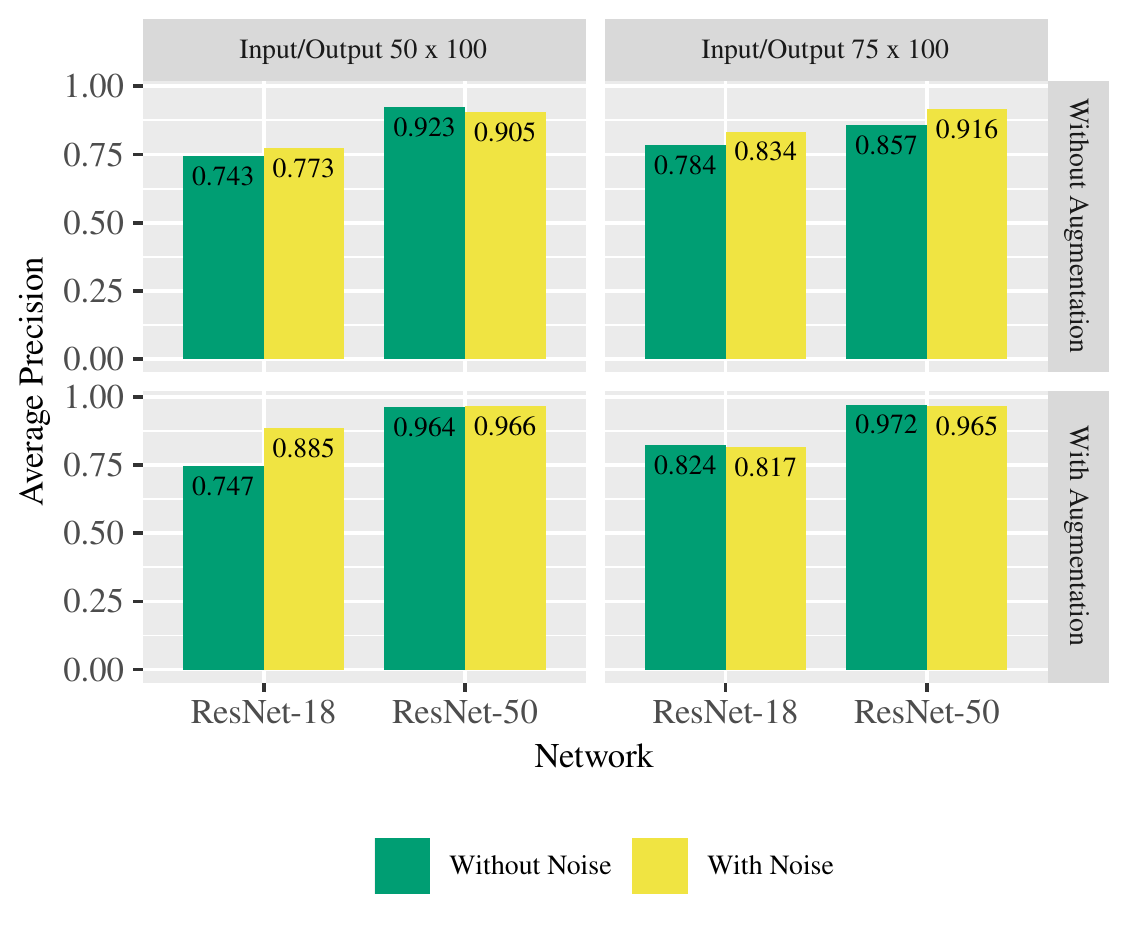}
	\caption{
		Results of our experiments on the figure skating dataset.
		We show the average precision for each model and directly compare the models with and without noise in the train dataset for the localizer.
		We also compare the results across two different feature extractors and assessors trained with and without augmentation.
	}
	\label{fig:figure_skating_experiments}
\end{figure}

\subsection{Figure Skating Experiments}

Following our experiments on the sheep dataset, we performed further experiments on the figure skating dataset, which we introduced in section~\ref{subsec:datasets}.
With these experiments, we wanted to achieve the following goals:
\begin{enumerate*}[label={(\arabic*)}]
	\item show that our approach works well across a range of different objects,
	\item show that it is easy to create a new dataset for training a model that performs very well, and
	\item show that our model is robust to noise in the localization training set.
\end{enumerate*}
Since we have no fully annotated dataset and also no access to the code of other weakly supervised object localization methods, we were not able to train a baseline model.
Instead, we trained a range of different models and examine the influence of different settings for training the model.
We trained all models on an input size of $224 \times 224$ pixels for the localizer, but we varied the following properties of the network:
\begin{enumerate*}[label={\arabic*.)}]
	\item the base model. We use ResNet-18 and a ResNet-50 that was pre-trained on Imagenet.
	\item The output size of the localizer / input size of the assessor ($50 \times 100$ pixels, or $75 \times 100$ pixels).
	\item The input dataset for the localizer. We either used the train dataset without noise, or the dataset with noise, as described in section~\ref{subsec:datasets}.
	\item We applied data augmentation on the train images of the assessor, by randomly adding horizontal flipping to \SI{50}{\percent} of the train images.
\end{enumerate*}

\begin{figure}[tbp]
	\centering
	\includegraphics[width=0.95\textwidth]{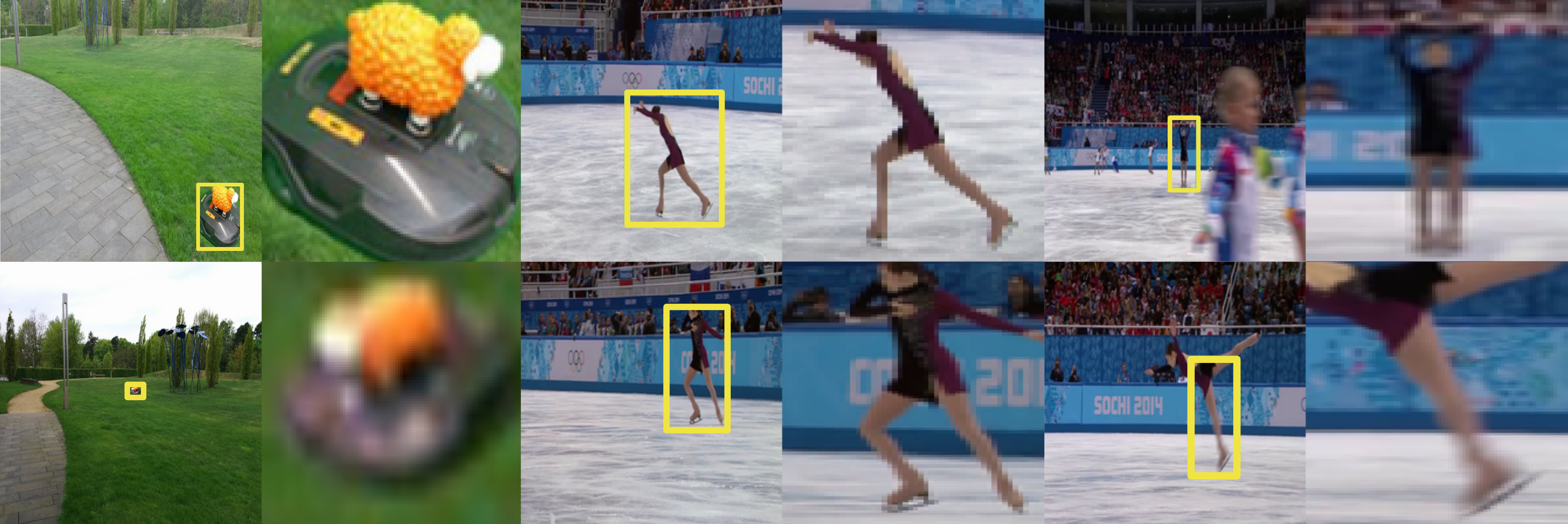}
	\caption{
		Samples from our test datasets, including localizations of our network from the sheep dataset (left column) and the figure skating dataset (middle and right column).
		The bottom-right image shows a failure case of our system, where our model is not able to accurately find the target object.
		The left part of each image is the input to the localizer and the right part the input to the assessor.
		The samples show that our model is capable of handling objects of different size, objects in different challenging positions, and also images that contain several distractors.
	}
	\label{fig:example_predictions}
\end{figure}

Figure~\ref{fig:figure_skating_experiments} shows the results of our experiments with the figure skating dataset, you can also find some samples of the dataset in figure~\ref{fig:example_predictions} in the middle and right columns.
In our supplementary material we provide a video showing the performance of our best model on the test video.
From the results, we can see that adding noise to our train dataset does not (necessarily) degrade the performance of the model.
Instead, it might even help the model to achieve better performance than before.
We argue that this due to the fact that the dataset with noise also contains images that show parts of a figure skater and the model uses this information to learn to localize the figure skater.
We also note that using data augmentation in the assessor leads to better results.
This shows that the training success of the localizer (student) depends on the assessor (teacher) and also shows us parallels to the behavior of human students and their teachers.

\section{Conclusion and Future Work}
\label{sec:discussion_and_future_work}

In this paper, we presented a novel approach for weakly supervised object detection, based on knowledge transfer between a teacher (assessor) and a student (localizer).
We evaluated our approach on two different datasets, showed that we can reach a performance that is on par to a fully supervised model, and that our approach is robust to noise in the training data.
Gathering training data for our approach is simple and does not need a lot of time.
For training the localizer, we can use real-world training data that has, for example, been extracted from a video.
We do not need any further annotation for training the localizer.
The train set for the assessor is an entirely synthetic dataset that can be created using a small amount of background and template images.
These properties should make it easy to use our approach for specialized object detection systems where it might be very expensive to fully annotate the training data.
Instead, taking a video and some pictures suffices for creating a dataset and training a very well performing model.
Our approach has some limitations that we want to address in our future work.
Our approach is currently not able to localize more than one object at the same time in an image.
We think that it is very important to further develop our idea and make it possible to detect multiple objects of different classes in the same image, in order to be able apply this approach to a broader class of problems.
Furthermore, we want to assess whether a model trained with our approach can be trained to work as a general object detector.
Preliminary experiments already showed that our model trained on the sheep dataset is able to localize other objects, such as a football.
Our model trained on the figure skating dataset works very well in localizing humans that are standing upright in an image, hence we see this as a good starting point to further investigate the generalization capabilities of our approach.


\bibliographystyle{splncs}
\bibliography{Remote}

\end{document}